\pdfoutput=1
\documentclass[letterpaper, 10 pt, conference]{ieeeconf}
\IEEEoverridecommandlockouts                

\usepackage[table, x11names]{xcolor}
\usepackage[colorlinks=true,linkcolor=blue,citecolor=blue,urlcolor=blue,pagebackref]{hyperref}
\usepackage{times} 
\usepackage{amsmath} 
\usepackage{amssymb}  
\usepackage{graphicx}
\usepackage{algorithm}
\usepackage{mathtools}
\usepackage{algorithm}
\usepackage{algorithmicx}
\usepackage[noend]{algpseudocode}
\usepackage[font=footnotesize]{subcaption}
\usepackage{enumitem}

\usepackage{cite}

\usepackage{balance}
\usepackage{multirow} 

\usepackage{booktabs}

\usepackage{glossaries-extra}
\setabbreviationstyle[acronym]{long-short}
\glssetcategoryattribute{acronym}{nohyperfirst}{true}

\makeglossaries

\newacronym{slam}{SLAM}{Simultaneous Localization and Mapping}
\newacronym{ba}{BA}{Bundle Adjustment}
\newacronym{sfm}{SfM}{Structure from Motion}
\newacronym{pgo}{PGO}{Pose-Graph Optimization}
\newacronym{vpr}{VPR}{Visual Place Recognition}
\newacronym{sgd}{SGD}{Stochastic Gradient Descent}
\newacronym{ils}{ILS}{Iterative Least-Squares}
\newacronym{gn}{GN}{Gauss-Newton}
\newacronym{lm}{LM}{Levenberg-Marquardt}
\newacronym{pcg}{PCG}{Preconditioned Conjugate Gradient}
\newacronym{map}{MAP}{Maximum-A-Posteriori}
\newacronym{gf}{GF}{Gaussian Filters}
\newacronym{pf}{PF}{Particle Filters}
\newacronym{sdp}{SDP}{Semi-Definite Programming}
\newacronym{vo}{VO}{Visual Odometry}
\newacronym{vio}{VIO}{Visual-Inertial Odometry}
\newacronym{lo}{LO}{LiDAR Odometry}
\newacronym{vlo}{VLO}{Visual-LiDAR Odometry}
\newacronym{imu}{IMU}{Inertial Measurement Unit}
\newacronym{ros}{ROS}{Robot Operating System}
\newacronym{ate}{ATE}{Absolute Trajectory Error}
\newacronym{rpe}{RPE}{Relative Pose Error}
\newacronym{vins}{VINS}{Visual INertial System}
\newacronym{pod}{POD}{Plain Old Data}
\newacronym{dpc}{DPC}{Dynamic Property Container}
\newacronym{boss}{BOSS}{Basic Object Serialization System}

\def\sota{state-of-the-art }
\def\slam{\gls{slam} }

\newacronym{kslam}{K-SLAM}{Keyframe-based \slam}

\def\secref#1{Sec.~\ref{#1}}
\def\figref#1{Fig.~\ref{#1}}
\def\tabref#1{Tab.~\ref{#1}}
\def\eqref#1{Eq.~(\ref{#1})}

\def\ie{\emph{i.e.}}
\def\eg{\emph{e.g.}}
\def\etal{\emph{et al.}}

\title{\LARGE \bf Plug-and-Play SLAM: A Unified\\  SLAM Architecture for
Modularity and Ease of Use}
\author{Mirco Colosi \hspace{15pt}%
  Irvin Aloise \hspace{15pt}%
  Tiziano Guadagnino \\
  Dominik Schlegel \hspace{15pt}%
  Bartolomeo Della Corte \hspace{15pt}%
  Kai O. Arras \hspace{15pt}%
  Giorgio Grisetti
\thanks{Mirco Colosi, Irvin Aloise, Tiziano Guadagnino, Dominik Schlegel,
  Bartolomeo Della Corte, Giorgio Grisetti are with the Department of Computer,
  Control, and Management Engineering  "Antonio Ruberti", Sapienza University of
  Rome, Rome, Italy, Email:\,\,{\tt\footnotesize{\{colosi, ialoise, guadagnino,
  schlegel, dellacorte, grisetti\}@diag.uniroma1.it}}.
  Mirco Colosi and Kai O. Arras are with Robert Bosch Corporate Research,
  Stuttgart, Germany.
  {\tt\footnotesize{\{mirco.colosi, kaioliver.arras\}@de.bosch.com}}.}%
\thanks{This work has been partially supported by Robert Bosch GmbH.}
}

\begin{document}
\maketitle
\thispagestyle{empty}
\pagestyle{empty}

\begin{abstract}
  Nowadays, \gls{slam} is considered by the Robotics community to be a
  mature field.  Currently, there are many open-source systems that are
  able to deliver fast and accurate estimation in typical real-world
  scenarios. Still, all these systems often provide an \mbox{ad-hoc}
  implementation that entailed to predefined sensor configurations. In
  this work, we tackle this issue, proposing a novel \gls{slam}
  architecture specifically designed to address heterogeneous sensor
  arrangement and to standardize \gls{slam} architecture.  Thanks to
  its modularity and to specific design patterns, the presented
  framework is easy to extend, enhancing code reuse and efficiency. Finally,
  adopting our solution, we conducted comparative experiments for a variety of
  sensor configurations, showing competitive results that confirms
  state-of-the-art performance.
\end{abstract}

\section{Introduction} \label{sec:intro}
\slam has become a mature research field with many applications
areas, ranging from autonomous vehicles to augmented reality. While there are
robust solutions for well posed use-cases - \eg~laser-based localization of
wheeled robots in planar environments~\cite{grisetti2007improved,hess2016real} - there are
scenarios in which either the
robot, the environment or the requirements are so challenging that a large
amount of further fundamental research is needed, as pointed out by Cadena
\etal~\cite{cadena2016past}.
%
%


In this context, multi-modal \gls{slam} can help to enhance the robustness of
the system, providing redundant information about the environment.
This could improve the
system performances in challenging scenarios or when a sensor is not suitable
to provide a specific feature - \eg~extracting colors from LiDAR data.
Multi-cues \gls{slam} has been explored over time by the research community and
many state-of-the-art systems support two or more sensors at the same time
- \eg~\gls{vlo} or \gls{vio}.
Still, the majority of these systems are meant to be used with a
\emph{predefined} combination of sensors. In this sense, they result difficult
to extend or to combine with other systems.

In this paper, we propose a custom \gls{slam} architecture that natively
supports heterogeneous sensors and aims at standardizing multi-modal \gls{slam}.
The architecture allows to mix-up different cues in a plug-and-play fashion
thanks to the isolation of the core \gls{slam} modules and, hence,
enhances code reuse and efficiency. In addition, exploiting specific
\gls{slam}-driven design patterns, our approach allows to embed
new cues even by simply editing a configuration file.
The entire architecture is oper-source and coded in
modern C++\footnote{Source code: \url{http://srrg.gitlab.io/srrg2.html}}.

\begin{figure}[t]
  \centering
  \includegraphics[width=0.99\columnwidth]{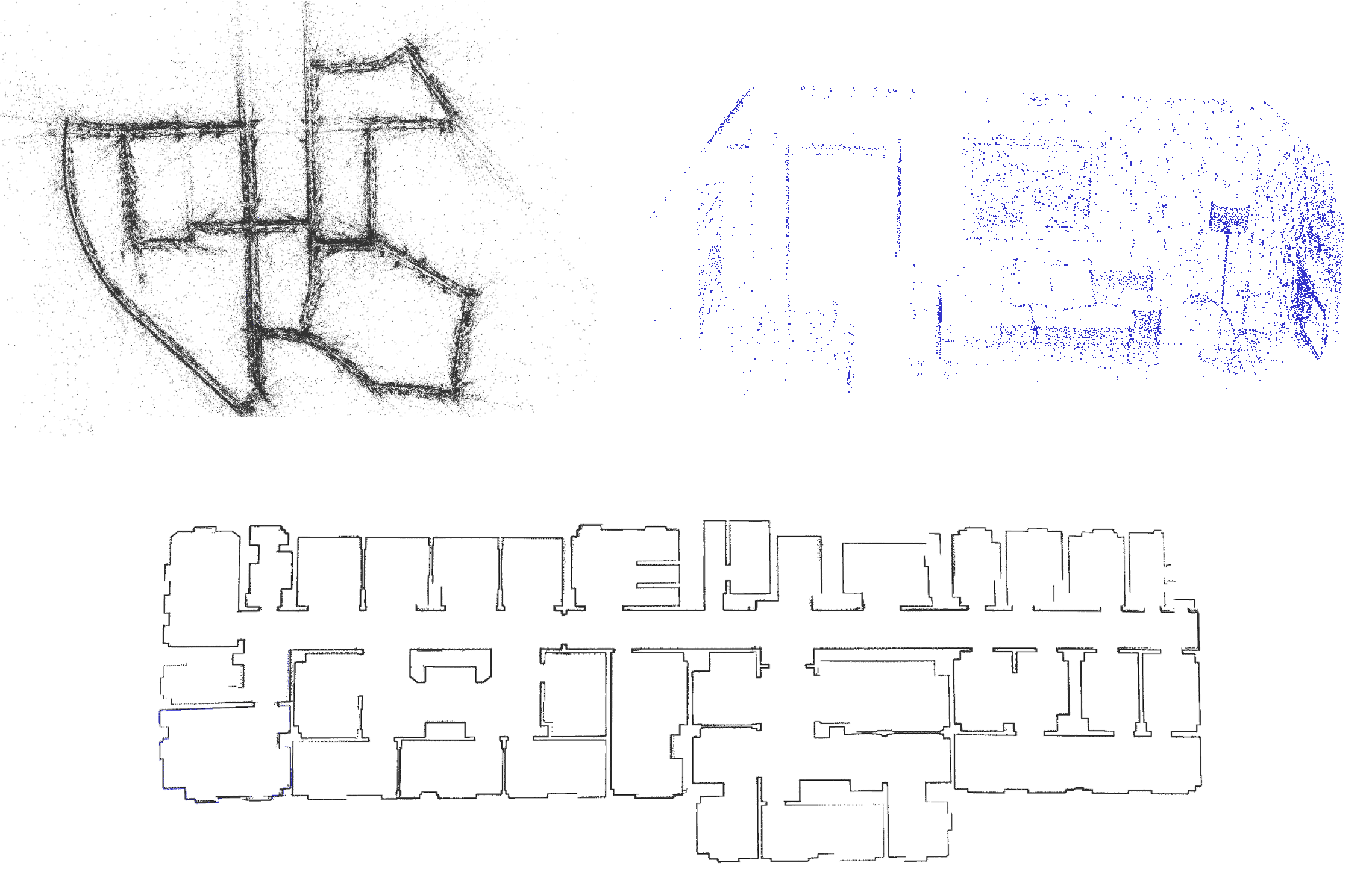}
  \caption{Result of 3 different SLAM pipelines, all embedded
    in our architecture. Top left represents \texttt{kitti-00},
    top right instead \texttt{icl-lr-0}. Bottom shows the map
    produced by the 2d-lidar pipeline on simulated data.}
  \label{fig:motivation}
\end{figure}

We validated our architecture using multiple 2D-LiDARs (in combination also
with wheel odometry), stereo and RGB-D cameras, resulting in outcomes
similar to ad-hoc state-of-the-art systems in all scenarios - as illustrated
in~\figref{fig:motivation}.

\vspace{5pt}

The remainder of this paper is organized as follows:
\secref{sec:related} presents a brief overview synopsis of multi-modal
\gls{slam} systems; in~\secref{sec:general-taxonomy}, instead, we
propose an overview of the generic building block of a \gls{slam}
system; in~\secref{sec:approach}, we show the multi-sensor our
architecture; finally, in~\secref{sec:experiments} we report the
results obtained with such an architecture with different sensors
configuration.

\section{Related Work}\label{sec:related}
In the context of \gls{slam}, sensor fusion indicates the capability
of a system of processing multiple cues at the same time. Multi-modal
\gls{slam} could dramatically improve the system performances in
various scenarios, especially when those are highly dynamic.  In the
past years the community addressed this topic, investigating ways of
integrating multiple cues in the same system.  A possible way of
exploiting multiple cues, is to have a main sensor and a supplementary
one. The latter is supports the system initialization or provides
specific cues such as the scale.
In the context of Visual-\gls{slam} this scenario is very common nowadays. Many
state-of-the-art system combine the use of a monocular camera and an \gls{imu}
to perform \slam
\cite{lynen2013robust,martinelli2014closed,mur2017visual,qin2018vins}.
In this sense, the \gls{imu} data is integrated over time~\cite{forster2015imu}
to produce a coarse estimate of the relative motion between two frames and to
infer the scale of the state. Similarly, in the work of
Pire~\etal~\cite{pire2017sptam}, the wheel odometry computed from
encoder readings, might be used to provide a prior in the registration of two
frames when using stereo cameras.
Lately, Rosinol~\etal~\cite{rosinol2019kimera} developed Kimera, a \gls{slam}
framework which combines camera images (either from a monocular or stereo setup)
together with \gls{imu} data to construct 3D metric-semantic maps.
In the context of LiDAR-based \gls{slam}, Zhang~\etal~\cite{zhang2015visual}
proposed to integrate range measurement and RGB data to estimate the sensor
motion. More specifically, the system initially computes the ego-motion through
\gls{vo} (high frequency but low fidelity) and then refines it exploiting
scan-matching based \gls{lo} (low frequency, high fidelity).
Newman~\etal~\cite{newman2006outdoor}, instead, used the additional cue coming
from RGB camera to compute loop-closure trough feature-based~\gls{vpr}.

In recent years, given the maturity of the \gls{slam} problem, the research
community started exploring the standardization and modularization of
\gls{slam} systems.
In this sense, closed-box architecture that can deal with specific sensors in a
pre-determined way leave room to dynamic multi-cues systems. Our work investigates along this research direction.
In this context, Schneider \etal~\cite{schneider2018maplab} proposed
\textit{maplab}, a framework to manage \gls{vio} in every aspect.
%
%
Therefore, \textit{maplab} is a Visual-Inertial Mapping and
Localization framework which unifies \sota \gls{vio}
implementations and map management or localization routines, allowing
multi-missions sessions. The authors offer various
off-the-shelf implementations of \sota algorithms and provide an architecture
that allows the user to integrate his own package in the framework. In
particular, \textit{maplab} allows to create a single open-loop map for every mission in
\gls{vio} mode, then stores the map and performs its refinement using efficient
off-line algorithms. As in our case, the user can interact with \textit{maplab}
through a console and provide it's own configuration. Still, this framework is
not intended to deal with multiple sensors other than a camera and \gls{imu}.

More recently, Blanco-Claraco proposed \textit{MOLA}~\cite{blanco19mola}, a
modular, flexible and fully extensible \slam architecture.  MOLA combines in a
single system multi-sensor capabilities and large map management, while being
completely customizable by the user. Examples of configuration parameters can
be the type of variable that represents the system state or the
\textit{back-end} in charge of performing global optimization.
%
%
MOLA has different types of independent sub-modules, each of which has a
specific role. In this sense, \textit{input} modules process raw sensor
readings, and act as data sources for \textit{front-end} modules. The latter
exploit standard \gls{slam} algorithm to create nodes and edges of the factor
graph, while the \textit{back-end} creates a unified interface to the
underlying global optimization framework - that can be chosen arbitrarily.
Finally, \textit{map-storage} modules
are in charge of storing and managing the map. These modules can also
dynamically serialize part of the total map to reduce memory usage.
%
MOLA gives the freedom to the user to completely define
the front-end module, who must implement some virtual functions for keyframe and
factor creation. In our work, instead, we detected some ``atomic" modules and
their connections to generate expected behaviors, resulting in a more
structured architecture that encourages the reuse of sub-modules.

Similarly, Labb{\'e}~\etal~\cite{labbe2019rtab} proposed a multi-sensor graph
\slam system called RTAB-Map.
The modularity is intrinsically granted by the use of \gls{ros}, by
which every processing module runs over a \gls{ros} node.
%
RTAB-Map was originally designed to be an appearance-based loop closure
detection approach~\cite{labbe2013appearance}, that was focused on memory
management to deal with long-term mapping sessions. Subsequently, RTAB-Map has
been highly expanded, resulting now in a Visual/Lidar \slam open-source library.
RTAB-Map can be used in two modalities. The first one, consists in a ``passive"
map manager, that takes as input odometry measurement - generated by some
external system - along with raw visual information. In this case, the system
maintains the map, detects loop closures and provides highly efficient memory
management.
In the "active" modality, RTAB-Map is able to generate itself the odometry
information, processing LiDAR or Visual data. In this sense, a great
variety of cues can be digested at the same time in a single framework. Still,
to extend the system, one has to completely develop a processing modules that
given raw sensor reading provides ego-motion estimation.

Most of the concept we adopt and extend have been previously explored in the
work of Colosi~\etal~\cite{colosi2019better}.
Here the authors defined a taxonomy of a generic graph-based \slam system. In
this definition, each presented component is responsible for a single task,
clearly defined by its input, outputs and mission. Though, the authors focused
single sensor scenarios.

The partition of a \slam system in components is also investigated in the survey
on Younes~\etal~\cite{younes2017keyframe}. In this work,
the authors design a generic \gls{kslam} flowchart made by several building
blocks. Furthermore,  they explain for each of them the expected
functionalities and the current \sota implementations available.
Even though, this work is only restricted to monocular camera systems. Still,
the idea behind the architecture is reasonably general and might be extended to
more generic graph-based \slam system, as we do in our work.

\section{Taxonomy of a Graph-Based SLAM System}\label{sec:general-taxonomy}
\begin{figure*}[!t]
  \centering
  \begin{subfigure}{\linewidth}
    \centering
    \includegraphics[width=0.8\linewidth]{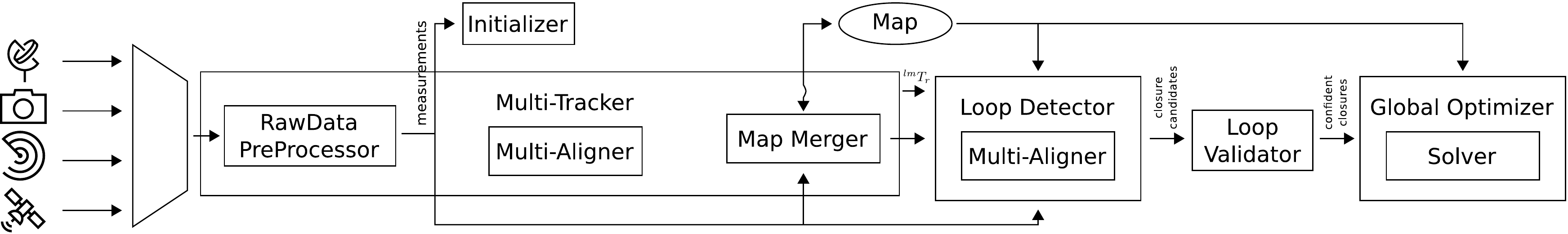}
    \subcaption{Multi-cues architecture.}
    \label{fig:architecture-blocks}
  \end{subfigure} \\ \vspace{5pt}
  \begin{subfigure}{0.47\linewidth}
    \centering
    \includegraphics[width=0.7\linewidth]{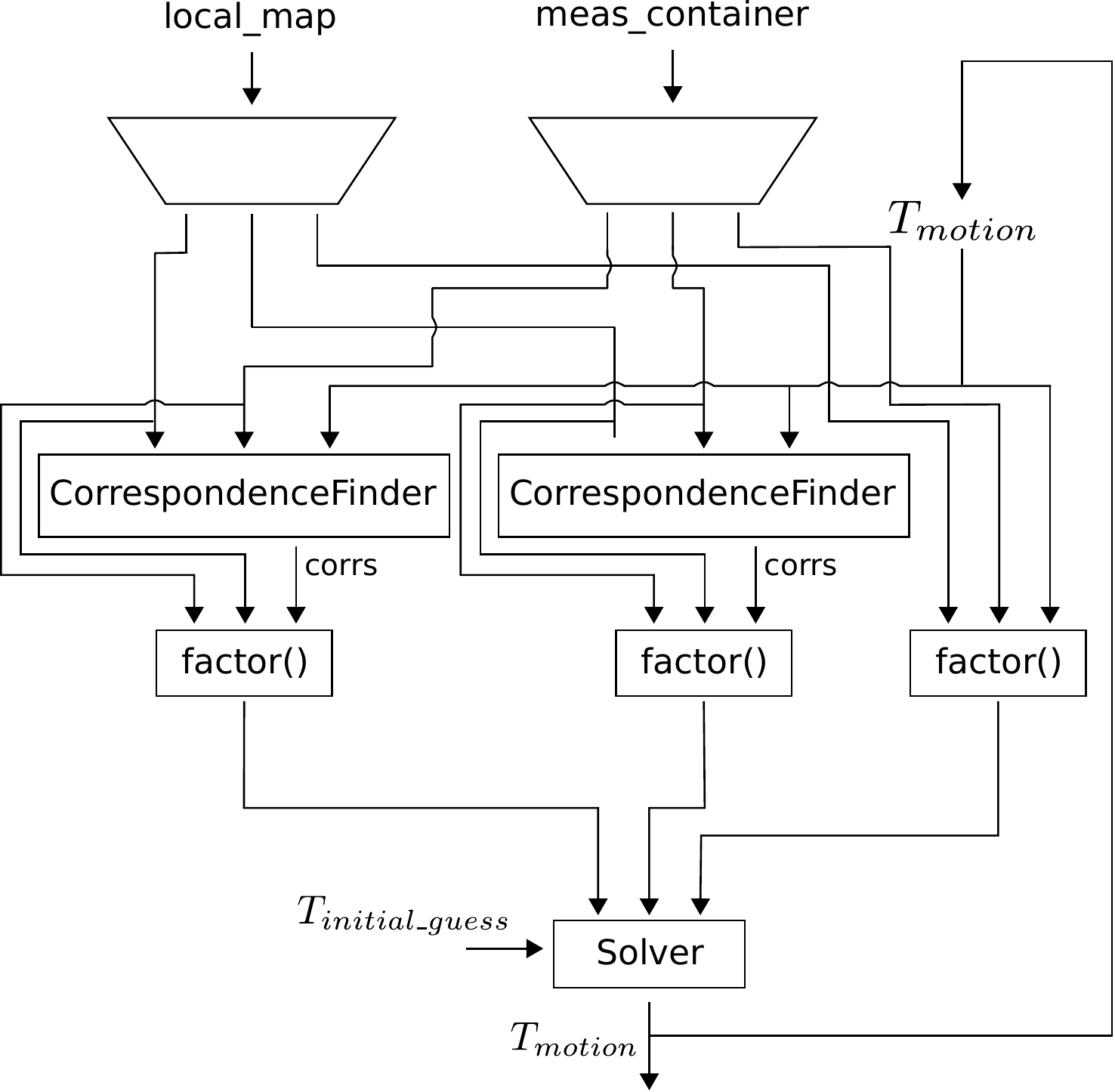}
    \subcaption{Multi-aligner.}
    \label{fig:architecture-aligner}
  \end{subfigure}
  \begin{subfigure}{0.47\linewidth}
    \centering
    \includegraphics[width=0.7\linewidth]{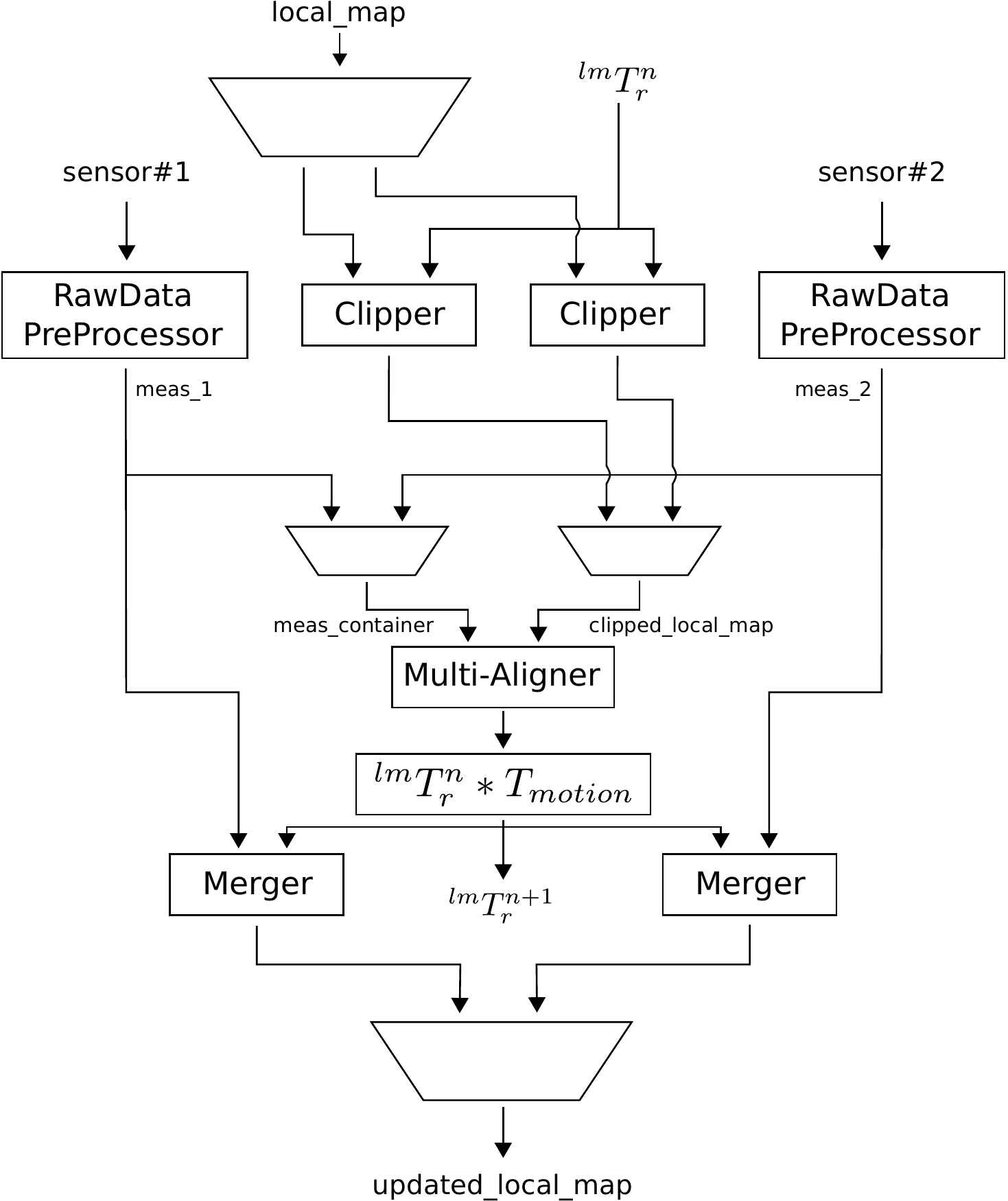}
    \subcaption{Multi-tracker.}
    \label{fig:architecture-tracker}
  \end{subfigure} \\ \vspace{5pt}
  \caption{Top image: blueprint of our multi-cues \gls{slam} architecture. Each
    sensor will contribute to populate the \emph{measurement property
      container}; this is fed into the Multi-Aligner to compute the relative
    motion of the robot; lately, the Multi-Tracker properly embeds each cue of
    the \emph{measurement property container} into the \emph{scene property
      container}; finally, the Graph-SLAM module arranges the local map into a
    factor graph, detects loop closures and optimizes the graph.
    \figref{fig:architecture-aligner} and~\figref{fig:architecture-tracker},
    instead, show a close-up of the secondary modules involved in the
    Multi-Aligner and Multi-Tracker respectively.
  }
  \label{fig:architecture}
\end{figure*}
In this section, we introduce the notions we endorse in this work. A modern
\slam system is generally composed by a group of modules which process a set of
shared data structures. Each processing module is in charge of performing
a relatively isolated task that takes the input data, processes them and
produces some output quantities.
Generally, the outcome of a \gls{slam} system can be represented trough a
factor graph~\cite{grisetti2010tutorial}. In this sense, the estimated
trajectory of the robot is represented through a pose-graph, a specialization
of a generic factor graph in which each variable represents a robot pose, while
factors encode spatial constraints between two poses. To avoid unbounded growth
of the factor graph, nodes are generally spawned according some kind of
heuristic - \eg~when the robot distance between the last variable's pose is
higher than a threshold. Variables in the graph, then, correspond to the pose
of the robot in these \emph{key-frames}.
Furthermore, one could ``attach'' to each key-frame, information about the
structure of the environment, represented by the \emph{landmarks}. Therefore,
each variable in the graph represents a rigid body, that we indicate as
\emph{local map}.

Colosi~\etal~\cite{colosi2019better} analyzed how a generic
single-sensor \gls{slam} system is composed. In the remaining of this
section, we review these concepts, while, in the next
section, we will extend them to multi-cues \gls{slam} systems.

\subsection{Core Modules}\label{sec:core-modules}
The workflow of a generic \gls{slam} system should %
i) process raw sensor's reading and generate data in a canonical format for the rest of system, %
ii) estimate the relative motion between two readings, %
iii) generate a trajectory and manage landmarks to create a consistent map and
finally %
iv) detect loop-closures and perform global optimization on the factor graph.
In this context the core modules involved can be summarized as follows:

\begin{description}[style=unboxed,leftmargin=0cm]
  \item \textsc{Raw Data Pre-processor:} as the name suggests, this module
  takes as input a raw sensor measurement and extracts suitable data-structures
  that can be used in the other modules. For example, given a RGB-D image, its
  output would consist in 3D visual-landmarks. We indicate with the term
  \emph{measurement} the output of such module.
  \item \textsc{Aligner:} this module would compute the relative motion between
  two sensor readings - or between a measurement and a local map. It is
  agnostic to the current system state since its
  only inputs are two entities (a fixed and a moving one) and possibly an
  initial guess of their offset.
  A possible implementation might exploit ICP to register two point-clouds -
  \eg~the one extracted from the last measurement
  and the previous one or the current local map.
  \item \textsc{Tracker:} it is in charge of managing and updating the current
  local map and generate a pose estimate of the traversing robot. Methods like
  \gls{vo} or scan-matching are typical instances of this module.
  \item \textsc{Graph-SLAM:} its task is to arrange local maps in a factor
  graph, detecting loop-closures and eventually trigger global optimization.
\end{description}

\subsection{Support Modules}\label{sec:support-modules}
Each core module owns and uses several other smaller sub-modules that could be
used to isolate specific tasks, enanching code reusabilty and modularity. The sub-modules combination in each core unit gives it a different
``flavor'', but does not affect its input-output or its role in the total
workflow. In the remaining, we illustrate the principal support modules of a
generic \gls{slam} system.

\begin{description}[style=unboxed,leftmargin=0cm]
  \item \textsc{Correspondence Finder:} given two compatible entities, computes
  the data association between them. Many implementation of this are possible,
  either based on their appearance~\cite{schlegel2018hbst,GalvezTRO12}
  or geometry~\cite{bentley1975multidimensional,dasarathy1991nearest}.
  \item \textsc{Merger:} its task is to incorporate new entities
    extracted from the current sensor reading in a local
    map. Different mapping approaches can be exploited also in this
    case.
  \item \textsc{Loop Detector and Validator:} These two modules are in
    charge of detecting loop-closures and run additional checks to
    reject false associations. Each accepted loop-closure will be
    turned to a new factor in the graph.
  \item \textsc{Global Optimizer:} it is in charge of performing
    non-linear optimization on the generated graph, to compute the
    variables configuration that best explains the factors. Note that,
    this could also be extended to pose-landmark configuration to
    accomplish map refinement.
\end{description}
Many other smaller components can be isolated in a \gls{slam} system. Examples
of them are the Map Clipper (generates a local view of the input map), Local
Map Splitter (triggers the creation of a new local map) or specific components
to bootstrap the \gls{slam} system - \eg~in monocular configuration to estimate
the 3D structure of the environment from a sequence of images.

\vspace{5pt}

Given this overview of the generic taxonomy of \gls{slam} systems, in the next
section we propose our design to seamlessly embed multiple heterogeneous cues
in a unified architecture.

\section{Multi-Sensor SLAM Architecture: Our Approach}\label{sec:approach}
The core idea behind our approach is that the computation performed with the
data coming from all the sensors contributes to estimate a single quantity: the
current robot's pose.
Therefore, referring to~\secref{sec:core-modules}, the only core modules
involved in this paradigm shift are Aligner and Tracker. Still, to allow
heterogeneous sensors to coexists in a unique architecture, one should also
define an appropriate data structure to represent multi-modal measurements and
local-maps. Note that, all the support modules are relatively agnostic to the
fact that the system works in a single or multi sensor configuration. Hence, in
the remaining of this section, we will first provide our design to store and
manage heterogeneous measurements and local-maps and then we will address the
changes in the aforementioned core modules.

\subsection{Multi-cues Data Structures}\label{sec:property-containers}
At the basis of our architecture, we have the concept of
\emph{Property}. A Property is an introspectable, serializable data
element, which is characterized data-type, value and by its name
within a containing structure the \gls{dpc}.  Properties can represent
basically anything, from \gls{pod} structures - \ie~a number or a
point cloud - or to entire modules - named \emph{Configurables}.
Thanks to the introspection, accessing at run time a specific Property
in a \gls{dpc}, requires just to know its name within.  Given this, we
can store different types of measurement in a \gls{dpc}, resulting in
what we call a \emph{measurement property container}. This will
contain the output of all Data Pre-processors currently instantiated
in the pipeline.  We can reuse the same machinery to store multi-modal
local map - here indicated as \emph{scene property container}.  In
this way, we are able to isolate different cues in a modular fashion,
while at the same time, we can provide a single input/output data
structure to the core modules.

\begin{table*}[!t]
  \centering
  \begin{tabular}{c>{\centering\arraybackslash}p{25mm}>{\centering\arraybackslash}p{25mm}>{\centering\arraybackslash}p{25mm}>{\centering\arraybackslash}p{25mm}}
    \toprule
    \multirow{2}{*}{\textsc{Approach}} &
    \textsc{stage-0} &
    \textsc{stage-1} &
    \textsc{stage-2} &
    \textsc{stage-3} \\
    {} & 
    $l = 43.191$ [m] &
    $l = 79.146$ [m] &
    $l = 480.741$ [m] &
    $l = 627.709$ [m] \\ \midrule
    \multirow{3}{*}{\textsc{srrg\_mapper2D}} &
    ATE = 0.258 [m] &
    ATE = 0.295 [m] &
    ATE = 0.287 [m] &
    ATE = 0.251 [m] \\
    {} & 
    RPE = 0.656 [m] &
    RPE = 0.777 [m] &
    RPE = 1.860 [m] &
    RPE = 1.529 [m] \\
    {} & 
    71.57 [Hz] &
    67.28 [Hz] &
    39.56 [Hz] &
    28.35 [Hz] \\ \midrule
    \multirow{3}{*}{\textsc{Our - Single LiDAR}} &
    ATE = \textbf{0.037} [m] &
    ATE = 0.041 [m] &
    ATE = 0.109 [m] &
    ATE = 0.110 [m] \\
    {} & 
    RPE = \textbf{0.059} [m] &
    RPE = \textbf{0.073} [m] &
    RPE = 0.298 [m] &
    RPE = 0.245 [m] \\
    {} & 
    \textbf{289.51} [Hz] &
    \textbf{278.07} [Hz] &
    \textbf{251.51} [Hz] &
    \textbf{189.58} [Hz] \\ \midrule
    \multirow{3}{*}{\textsc{Our - Double LiDAR}} &
    ATE = 0.047 [m] &
    ATE = \textbf{0.032} [m] &
    ATE = \textbf{0.072} [m] &
    ATE = \textbf{0.047} [m] \\
    {} & 
    RPE = 0.084 [m] &
    RPE = 0.077 [m] &
    RPE = \textbf{0.256} [m] &
    RPE = \textbf{0.127} [m] \\
    {} & 
    153.88 [Hz] &
    149.94 [Hz] &
    144.42 [Hz] &
    125.65 [Hz] \\ \bottomrule
  \end{tabular}
  \caption{Comparison of \gls{ate} and \gls{rpe}
    between~\cite{lazaro2018efficient} and our approach. All
    approaches exploit
    also wheel odometry together with LiDAR data. For each approach, the last
    row reports the mean processing frame-rate.}
  \label{tab:laser-results}
\end{table*}

\subsection{Multi-cues Core Modules}\label{sec:multi-modules}
Once addressed how to store dynamic and multi-modal data, we tackle in
this section the problem of processing them.  In this scenario, the
foundation of our approach relies on the concept of \emph{slice}.  A
\emph{slice} is a partial processing module, in charge of treating a
specific cue of the architecture.  Therefore, we can add to the
Aligner and Tracker the capability of having multiple \emph{slices}
designed addresses a specific sensor reading type.

More in detail, a Multi-Cue Aligner, called Multi-Aligner for brevity,
is composed by a single \gls{ils} solver and a set of
\emph{aligner-slices}. The former is in charge of optimizing the
registration graph to compute the robot motion, while the latter
produces factors for the different sensors to be fed into the
registration graph.  As in the single-sensor case, the variable to be
estimated remains only the robot relative motion, still, all factors
concerning different sensors will concurrently affect the estimate.

The same reasoning is applied to the Multi-Cue Tracker - shortened as
Multi-Tracker. It is composed by a single local map - stored through a
\gls{dpc} - and multiple \emph{tracker-slices}. Given a \emph{measurement
property container} and a \emph{scene property container} as local map, each
tracker-slice will work to process and manage a single cue from them.
\figref{fig:architecture} provides a schematic illustration of the entire
multi-cues workflow.

To give a more practical insight of the architecture, we can address the case
of having a 2D-LiDAR pipeline with two rangefinders - one front-facing and the
other rear-facing. The system will have two separate Data Pre-processors.
Accordingly, the \emph{measurement property container} is composed by two
Properties - relative to the point cloud extracted from the two sensors - and
the local map is a \gls{dpc} with two point-cloud too.
The Multi-Aligner is composed by 2 \emph{slices}, one for
the front rangefinder and the other for the rear one. Both of them will compute
the data association between its current measurement and the local-map.
Each of the two \emph{slices} will expose a set of constraints coming from the
association, that will be used by the Multi-Aligner to estimate the motion. The
same applies to the Multi-Tracker. Both \emph{slices} will take a point cloud
from the \emph{measurement property container} and will integrate it in the
relative point cloud of the local map.

Note that, thanks to this design, every other module in the
architecture remains agnostic to the number of sensor involved in the pipeline.
This means that one can potentially mix-up different modules in a plug and play
fashion, without the need of further modification to the architecture.

\subsection{Complementary Features}
Beside the \gls{slam} related benefits of such architecture, the
design pattern that we employed brings advantages to other
contexts.  In this sense, thanks to the native serialization of each
Property, we are able to automatically store the graph and the
local map on disk. This is carried on by our custom-built library,
that supports format-independent serialization of arbitrary data
structures - called \gls{boss}.  Furthermore, each processing module -
named \emph{Configurable} - exposes its parameters through
Properties. This allows us to use \gls{boss} to write a module
configuration automatically on disk. Note that, since also
Configurables can be stored in Properties, we are able to
\emph{instanciate} an entire pipeline reading from the configuration
file.  Even if such file is written in a human-readable format - based
on JSON - editing a complex configuration by hand might result
difficult. Still, thanks to the native introspection of the Property,
we developed a graphical editor, to edit \gls{boss} configuration file
on-the-go.  We exploit the same Property features to provide the user
with a shell to load and run configurations. Finally, Configurable
entities allow to expose module \emph{actions} that can be triggered
at runtime. For example, one can pause the pipeline and save the
factor graph on disk at runtime, simply typing commands in our shell.
Note that, the proposed architecture embeds our custom optimization framework
as back-end~\cite{grisetti2020squares} that shares the same core
functionalities - \eg~configuration management, serialization library.
In this sense, also the graph-optimization module remains consistent with the
rest of the architecture.

Furthermore, we provide also a unified viewing system based on OpenGL,
that decouples processing and viewing.  Using this API, one can either
run the SLAM pipeline and its visualization on the same machine in two
separate threads or run a \gls{slam} pipeline on a cheap embedded
system and stream the visual information to a more powerful machine
that will act as a passive rendering viewport.  Switching between these
two modalities does not require to change the code.
Currently, the multi-process viewing system relies on the
\gls{ros} communication infrastructure to share data.

\section{Experiments}\label{sec:experiments}
In this section we provide both qualitative and quantitative results obtained
though the \emph{instanciation} of different \gls{slam} pipelines embedded in
our architecture. In this sense, the purpose of this section is to show that
completely different pipelines are able not only to coexist together, but they
also achieve competitive results, and, thus, there is no apparent negative
impact on the systems' performances using our architecture.
Thanks to the modular design of the proposed approach, the system
natively allows to mix heterogeneous sensor - \eg~LiDAR-2D and RGBD - in a
unique multi-cues pipeline.
In the remaining of this section, we will show the results obtained with
LiDAR-2D, Stereo and RGB-D \gls{slam} pipeline instantiations respectively.
All the experiments have been performed on laptop with Ubuntu 18.04 and GCC 7,
equipped with an Intel Core i7-7700HQ @ 2.80 GHz and 16GB of RAM. Note that,
all the processing in our architecture is \emph{single threaded}.
\begin{figure*}[!t]
  \centering
  \begin{subfigure}{0.49\linewidth}
    \centering
    \includegraphics[width=0.8\columnwidth]{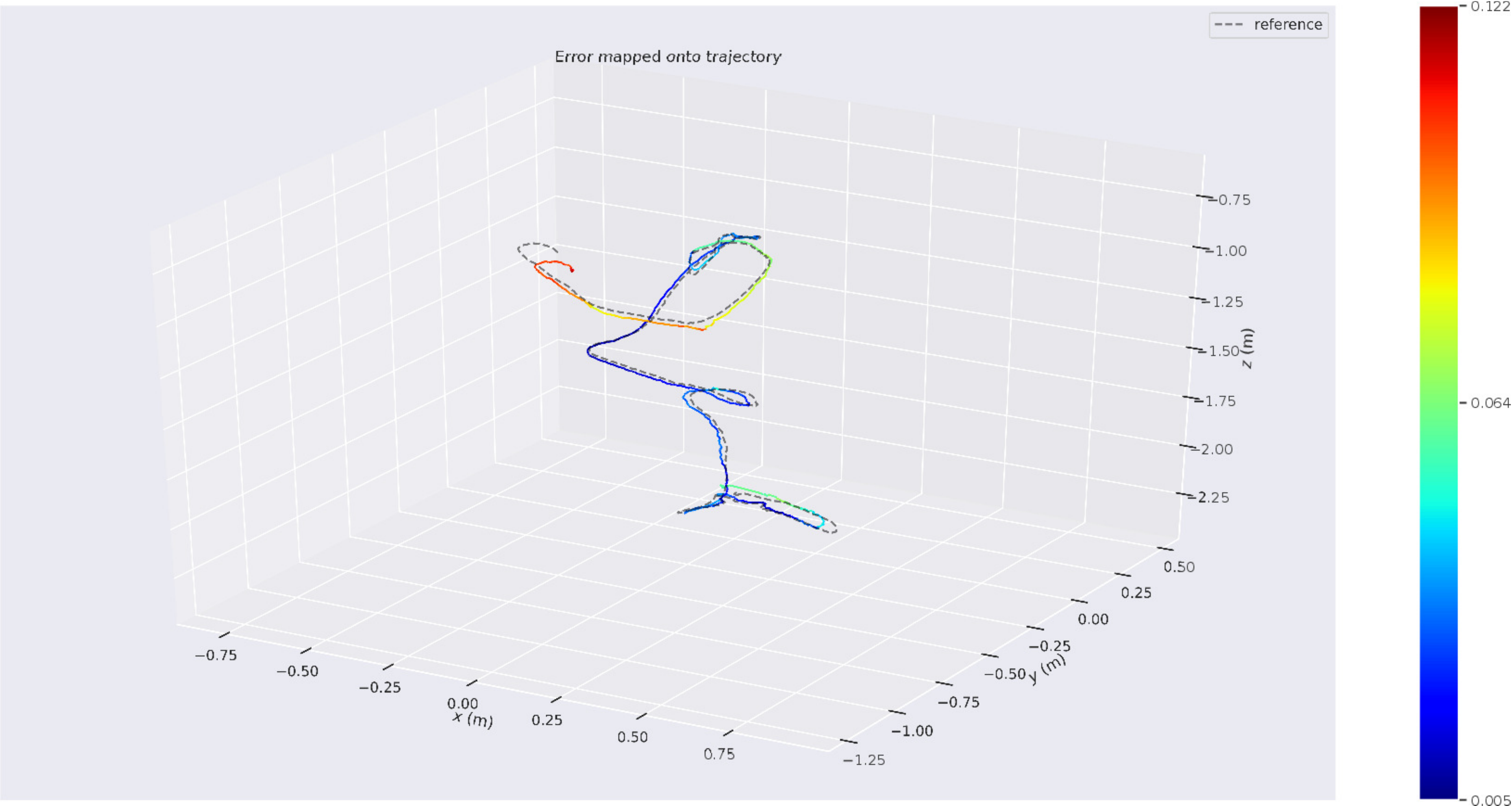}
    \subcaption{ICL \texttt{lr-0}: ProSLAM.}
    \label{fig:visual-slam-trajectories-lr0-proslam}
  \end{subfigure}
  \begin{subfigure}{0.49\linewidth}
    \centering
    \includegraphics[width=0.8\columnwidth]{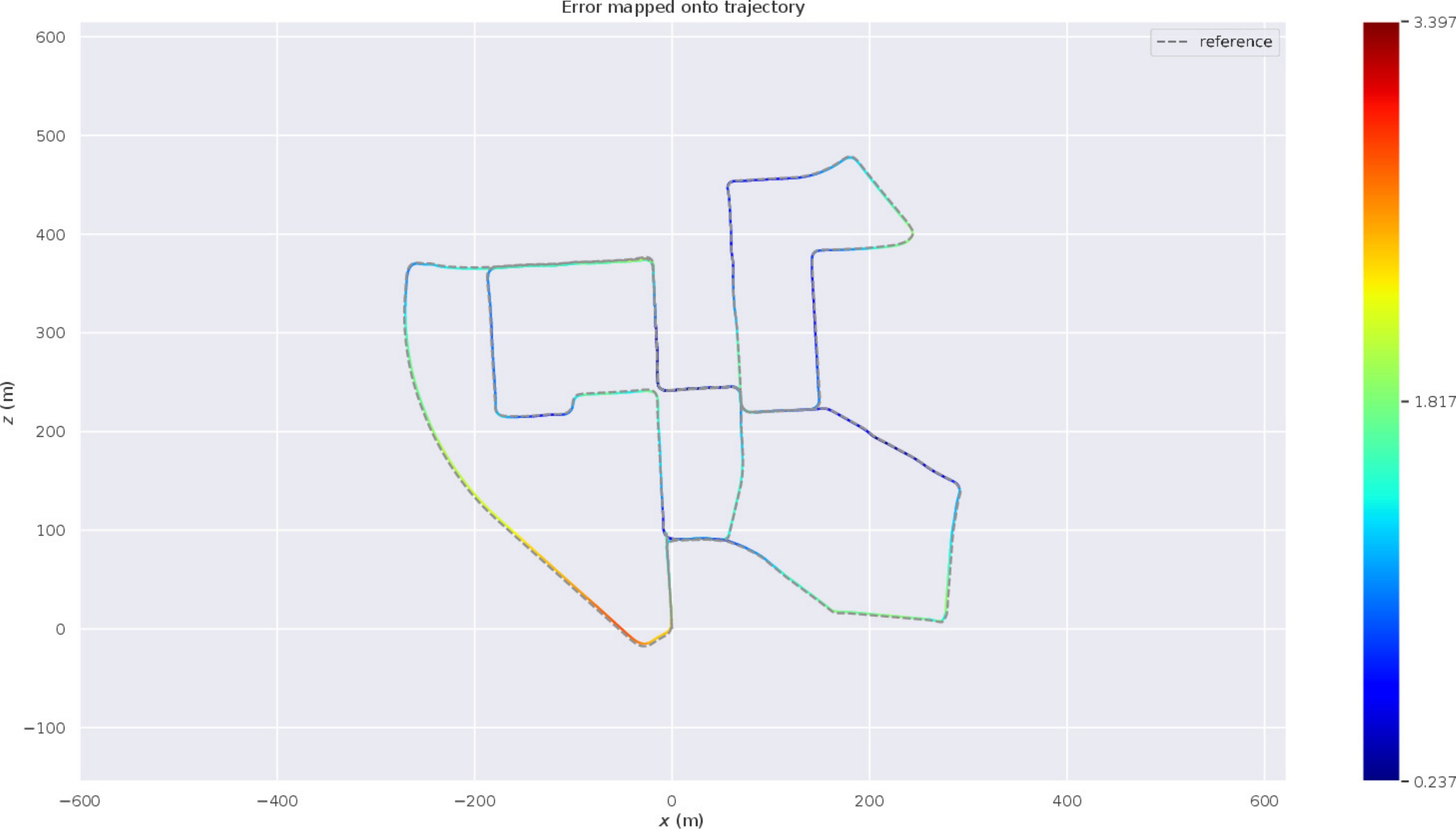}
    \subcaption{KITTI \texttt{00}: ProSLAM.}
    \label{fig:visual-slam-trajectories-kitti00-proslam}
  \end{subfigure} \\ \vspace{5pt}
  \begin{subfigure}{0.49\linewidth}
    \centering
    \includegraphics[width=0.8\columnwidth]{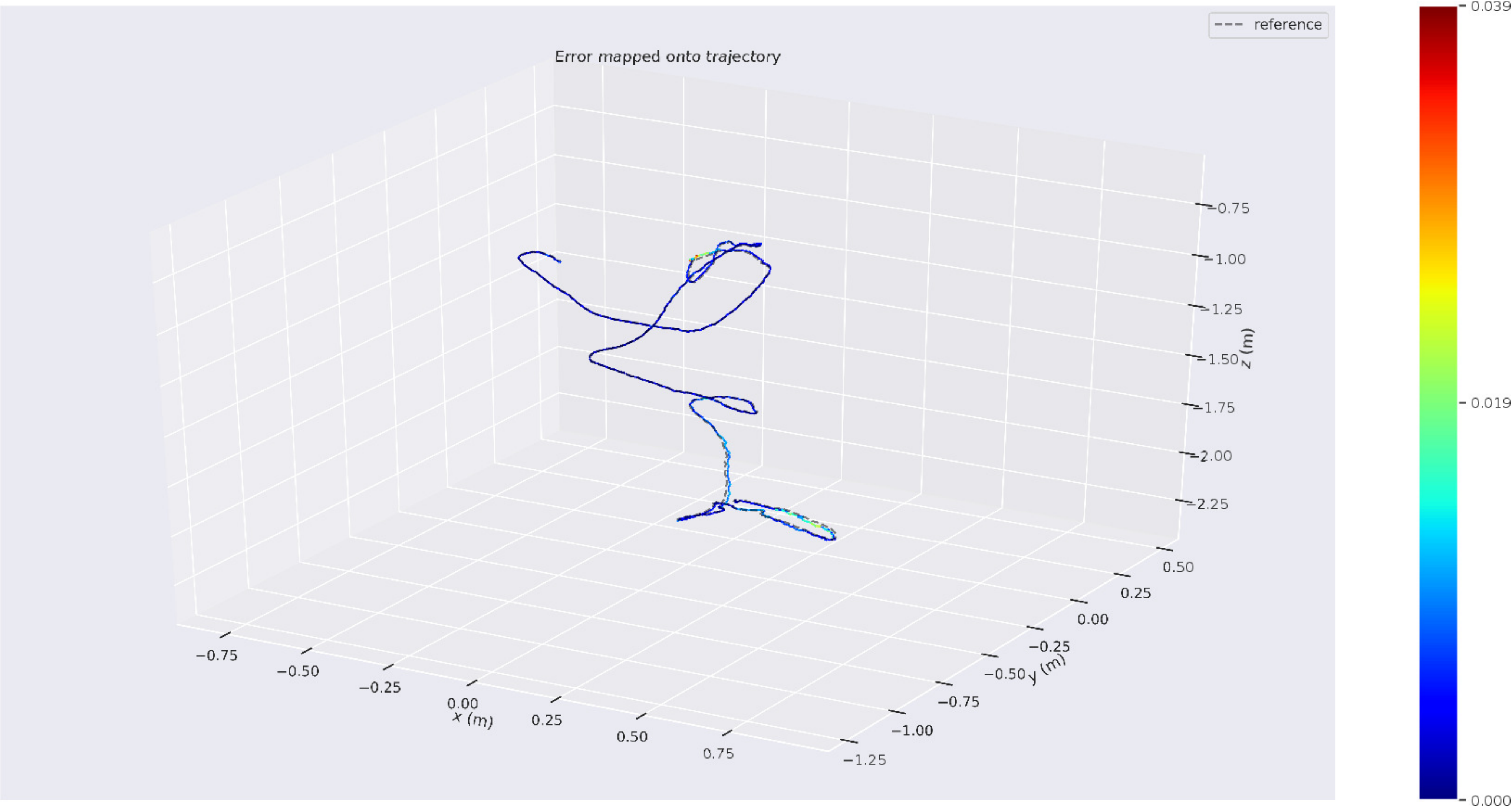}
    \subcaption{ICL \texttt{lr-0}: ORB-SLAM2.}
    \label{fig:visual-slam-trajectories-lr0-orb2}
  \end{subfigure}
  \begin{subfigure}{0.49\linewidth}
    \centering
    \includegraphics[width=0.8\columnwidth]{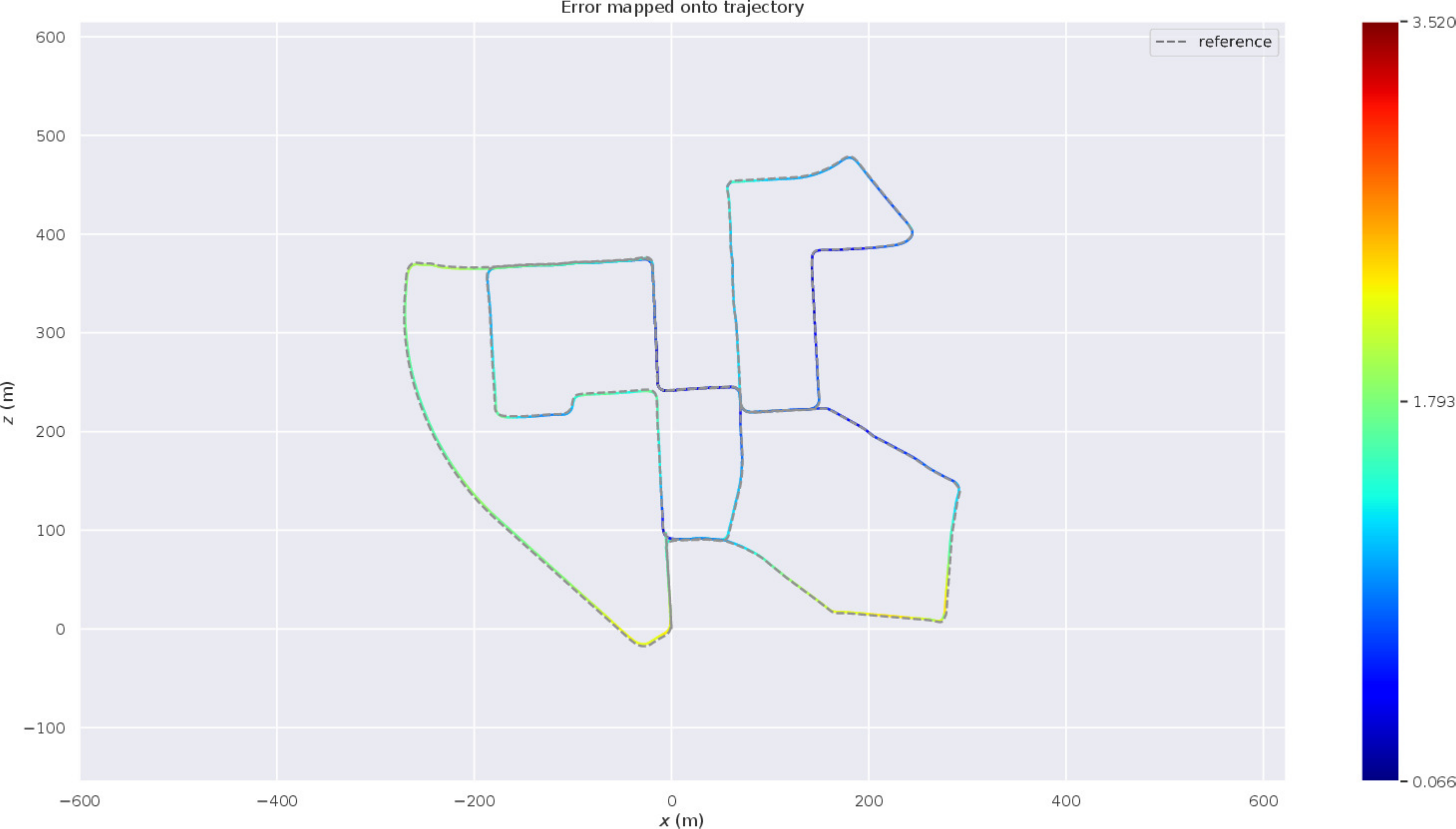}
    \subcaption{KITTI \texttt{00}: ORB-SLAM2.}
    \label{fig:visual-slam-trajectories-kitti00-orb2}
  \end{subfigure} \\ \vspace{5pt}
  \begin{subfigure}{0.49\linewidth}
    \centering
    \includegraphics[width=0.8\columnwidth]{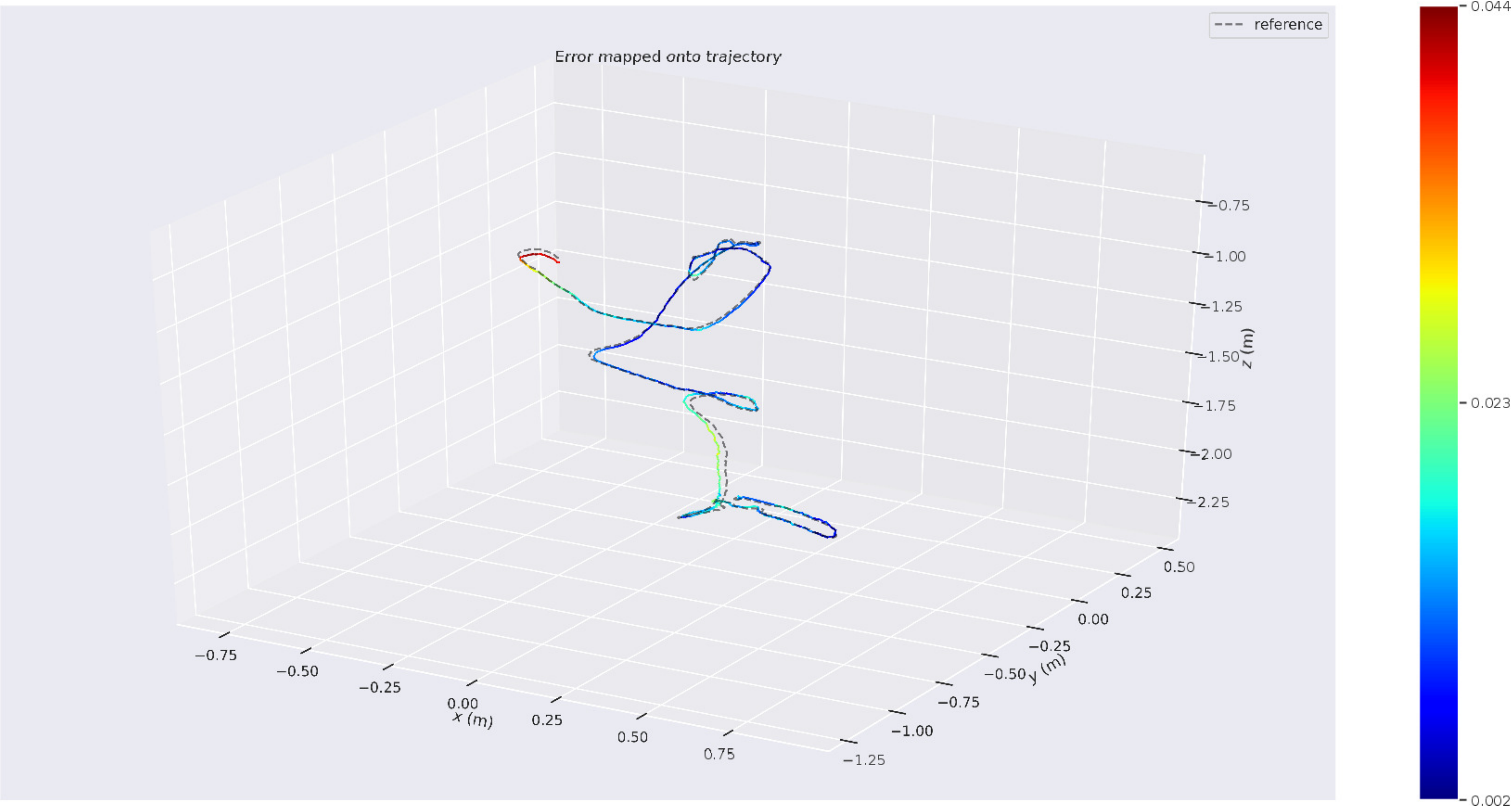}
    \subcaption{ICL \texttt{lr-0}: our.}
    \label{fig:visual-slam-trajectories-lr0-proslam2}
  \end{subfigure}
  \begin{subfigure}{0.49\linewidth}
    \centering
    \includegraphics[width=0.8\columnwidth]{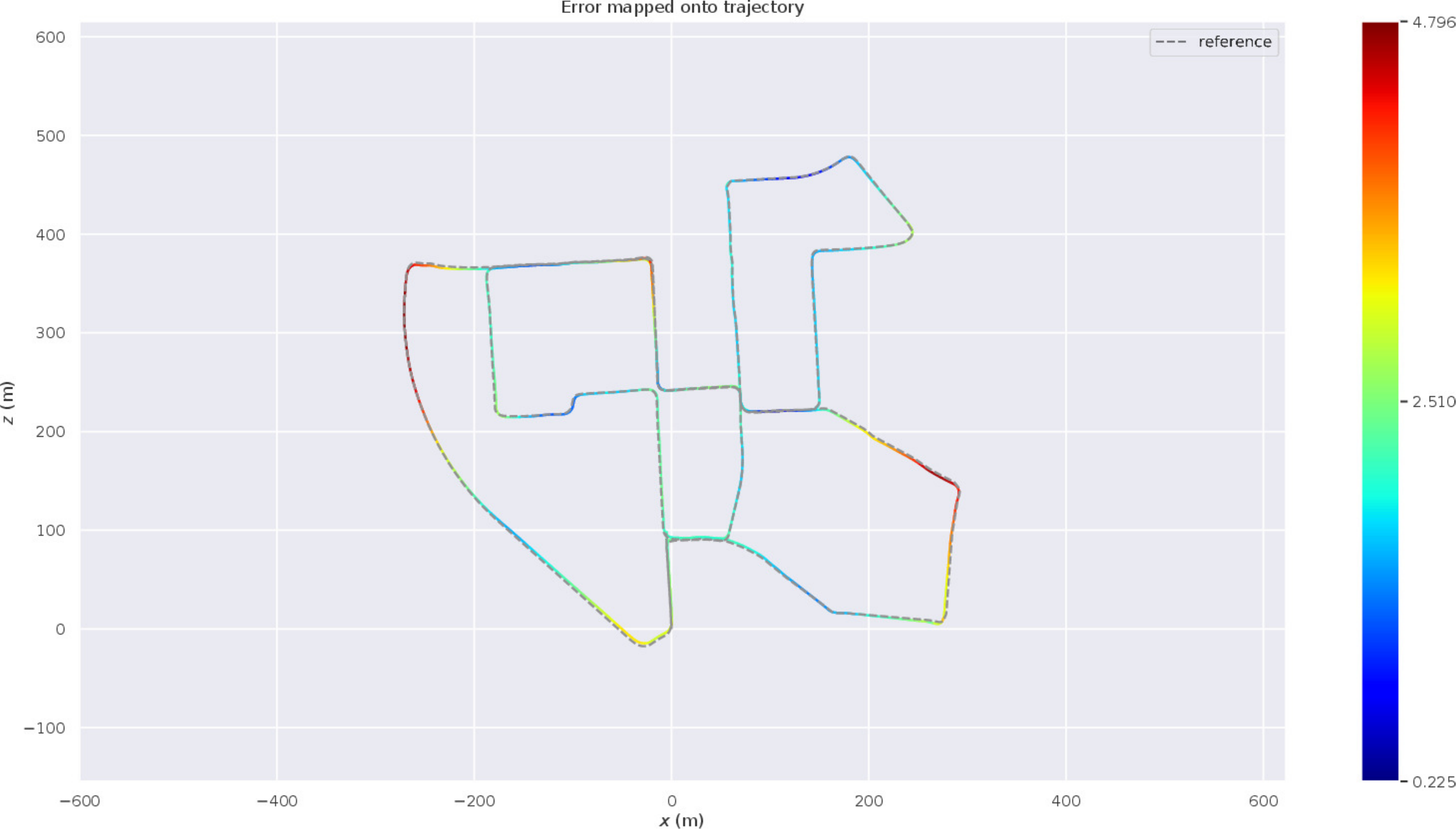}
    \subcaption{KITTI \texttt{00}: our.}
    \label{fig:visual-slam-trajectories-kitti00-proslam2}
  \end{subfigure}
  \caption{Trajectory comparison between ProSLAM, ORB-SLAM2 and our
    approach in the context of Visual-\gls{slam} - stereo and RGB-D.}
  \label{fig:visual-slam-trajectories}
\end{figure*}

\subsection{LiDAR-2D}\label{sec:laser-slam}
The 2D laser rangefinder is a very common sensor in \gls{slam} and it has been
employed since many years. Nowadays, it can be considered a cheap sensor and,
thus, it is now spreading in consumer Robotics. Many open-source systems are
available in this context
\cite{grisetti2007improved,lazaro2018efficient,hess2016real},
however, the majority of those are designed to be used with only one sensor cue and
their extension might require a lot of effort.
Our approach, instead, can be arranged to work with multiple
rangefinders and/or in combination with wheel odometry by simply editing a
configuration file.

To check the performance of our pipeline, we used simulated data, gathered
using \gls{ros} Stage~\cite{vaughan2008massively}. We recorded
multiple sessions with different path lengths.
The simulated differential-drive platform is equipped with
2 laser rangefinders (front and rear facing, horizontally mounted) and wheel
encoders that provide wheel odometry - all streaming data at 10 Hz.
In~\tabref{tab:laser-results} we reported a comparison
between the approach of Lazaro~\etal~\cite{lazaro2018efficient} - referred as
\texttt{srrg\_mapper2d} - and our approach on the different sessions.
Both the~\gls{ate} and~\gls{rpe} are lower than the one obtained with the
reference system. Using the information coming from the second
rangefinder increases the accuracy as expected, while system speed remains more
than 10 times faster than the platform sensor frame-rate.
\begin{table}[!t]
  \centering
  \begin{tabular}{c>{\centering\arraybackslash}p{30mm}>{\centering\arraybackslash}p{30mm}}
    \toprule
    \multirow{2}{*}{\textsc{Approach}} &
    \textsc{kitti-00} &
    \textsc{icl-lr-0} \\
    {} & 
    $l = 3724.187$ [m] &
    $l = 6.534$ [m] \\ \midrule
    \multirow{3}{*}{\textsc{ProSLAM}} &
    ATE = 1.378 [m] &
    ATE = 0.049 [m] \\
    {} & 
    RPE = 0.041 [m] &
    RPE = 0.003 [m] \\
    {} & 
    \textbf{49.30} [Hz] &
    \textbf{62.97} [Hz] \\ \midrule
    \multirow{3}{*}{\textsc{ORB-SLAM2}} &
    ATE = \textbf{1.336} [m] &
    ATE = \textbf{0.007} [m] \\
    {} & 
    RPE = \textbf{0.029} [m] &
    RPE = 0.004 [m] \\
    {} & 
    11.29 [Hz] (4 threads)&
    45.28 [Hz] (4 threads)\\ \midrule
    \multirow{3}{*}{\textsc{Our}} &
    ATE = 2.469 [m] &
    ATE = 0.016 [m] \\
    {} & 
    RPE = 0.037 [m] &
    RPE = \textbf{0.002} [m] \\
    {} & 
    43.88 [Hz] &
    49.98 [Hz] \\ \bottomrule
  \end{tabular}
  \caption{Comparison of \gls{ate} and \gls{rpe}
    between ProSLAM~\cite{schlegel2018proslam}, ORB-SLAM2~\cite{mur2017orb} and our approach. For each approach, the last
    row reports the mean processing frame-rate.}
  \label{tab:visual-results}
\end{table}

\subsection{Visual SLAM: RGB-D and Stereo}\label{sec:visual-slam}
Cameras represents one of the most employed sensor in the context of
\gls{slam}. Nowadays, stereo configurations and depth cameras (IR and ToF) are
becoming always more used in a great variety of domains, from Robotics to
consumer electronic.
In the proposed architecture, we addressed both stereo and RGB-D data in a
single-cue fashion. Still, in the near future we are expecting to enable
multi-cues pipelines as in~\secref{sec:laser-slam}.
To evaluate the performances of our pipelines we compared the results obtained
on the KITTI dataset~\cite{geiger2013vision} and on the ICL-NUIM
dataset~\cite{handa2014icl} with the state-of-the-art system
ProSLAM~\cite{schlegel2018proslam} and ORB-SLAM2~\cite{mur2017orb}.
In~\tabref{tab:visual-results} we reported the values of the \gls{ate} and
\gls{rpe} on the considered sequences, while
in~\figref{fig:visual-slam-trajectories} we
report a plot of the trajectories computed with each system.
The comparison confirms that our architecture is able to achieve results
comparable with other more mature state-of-the-art systems.

\section{Conclusions}\label{sec:conclusions}
In this paper we presented a novel architecture that aims
to standardize multi-sensor \gls{slam}. To achieve this goal, we
firstly analyzed the recurrent patterns in the context of \gls{slam}
systems, generating a taxonomy of sub-modules. Then, we
exploited such taxonomy to provide the user the ability to easily
integrate heterogeneous sensors in a single unified pipeline.
The paper exposes the design patterns and the data-structures
used in our implementation, presenting a modular and easy-to-extend
architecture. In our opinion, the latter could also have a major educational
impact.

Finally, we conducted a set of comparative experiments to show that
our architecture has no drawbacks on the accuracy of the
estimation nor on the runtime performances. In fact, even if pure
benchmarking is out of the scope of this work, the results obtained
are comparable with ad-hoc state-of-the-art implementations - in the
context of 2D-LiDAR, RGB-D and Stereo \gls{slam}.



\balance

\end{document}